\documentclass{ieeetj}
\usepackage{cite}

\usepackage{amsmath,amssymb,amsfonts}
\usepackage{algorithmic}
\usepackage{graphicx,color}
\usepackage{textcomp}
\usepackage{xcolor}
\usepackage{subcaption}
\usepackage{hyperref}
\hypersetup{hidelinks=true}
\usepackage{algorithm,algorithmic}
 \usepackage{url}
\usepackage{cancel}
\usepackage{amsmath,amssymb,amsfonts}
\DeclareMathOperator*{\argmax}{arg\,max}
\usepackage{booktabs}
\usepackage{textcomp}
\usepackage{pifont}
\usepackage{tikz}
\newcommand*\circled[1]{\tikz[baseline=(char.base)]{
            \node[shape=circle,draw,inner sep=0.4pt] (char) {#1};}}
\newcommand{\xmark}{\ding{55}}%

\def\BibTeX{{\rm B\kern-.05em{\sc i\kern-.025em b}\kern-.08em
    T\kern-.1667em\lower.7ex\hbox{E}\kern-.125emX}}
\AtBeginDocument{\definecolor{tmlcncolor}{cmyk}{0.93,0.59,0.15,0.02}\definecolor{NavyBlue}{RGB}{0,86,125}}

\def\OJlogo{\vspace{-4pt}$<$Society logo(s) and publication title will appear here.$>$}
\def\seclogo{\vspace{10pt}$<$Society logo(s) and publication title will appear here.$>$}

\def\authorrefmark#1{\ensuremath{^{\textbf{#1}}}}

\begin{document}
\bstctlcite{IEEEexample:BSTcontrol}
\receiveddate{XX Month, XXXX}
\reviseddate{XX Month, XXXX}
\accepteddate{XX Month, XXXX}
\publisheddate{XX Month, XXXX}
\currentdate{XX Month, XXXX}
\doiinfo{XXXX.2022.1234567}

\markboth{}{Arora {et al.}: Chain-of-Thought Reasoning in Streaming Full-Duplex End-to-End Spoken Dialogue Systems}

\title{Chain-of-Thought Reasoning in Streaming Full-Duplex End-to-End Spoken Dialogue Systems}

\author{Siddhant Arora\authorrefmark{1}, Jinchuan Tian\authorrefmark{1}, Hayato Futami\authorrefmark{2}, Jiatong Shi\authorrefmark{1}, \\ Yosuke Kashiwagi\authorrefmark{2}, Emiru Tsunoo\authorrefmark{2}, Shinji Watanabe\authorrefmark{1}}
\affil{Language Technologies Institute, Carnegie Mellon University, USA}
\affil{Sony Group Corporation, Japan}
\corresp{Corresponding author: Siddhant Arora (email: siddhana@andrew.cmu.edu).}

\begin{abstract}
Most end-to-end (E2E) spoken dialogue systems (SDS) rely on voice activity detection (VAD) for turn-taking, but VAD fails to distinguish between pauses and turn completions. 
Duplex SDS models address this by predicting output continuously, including silence tokens, thus removing the need for explicit VAD. However, they often have complex dual-channel architecture and lag behind cascaded models in semantic reasoning.
To overcome these challenges, we propose \emph{SCoT}: a Streaming Chain-of-Thought (CoT) framework for Duplex SDS, alternating between processing fixed-duration user input and generating responses in a blockwise manner. Using frame-level alignments, we create intermediate targets—aligned user transcripts and system responses—for each block. Experiments show that our approach produces more coherent and interpretable responses than existing duplex methods while supporting lower-latency and overlapping interactions compared to turn-by-turn systems.
\end{abstract}


\maketitle

\section{INTRODUCTION}
\label{sec:intro}
Spoken Dialogue Systems (SDS)~\cite{jokinen2009spoken,breazeal2008social} aim to build natural and interactive conversation with the end user.
SDS take a continuous audio stream as input from the user and generate a corresponding spoken response. They power everyday technologies — intelligent assistants like Alexa and Siri, and interactive voice response systems in customer service. More recently, there’s been growing interest in bringing these capabilities\footnote{\url{https://openai.com/index/hello-gpt-4o/}, \url{https://deepmind.google/technologies/gemini/}} to mobile phones and wearable devices. As these systems become more common, the need for robust, scalable, and generalizable SDS becomes critical.

SDS have traditionally operated in a turn-by-turn manner~\cite{glass1999challenges,raux2005let,Qwen2.5-Omni}, where the system alternates between two distinct modes: listening and speaking.
The segmentation of turns can be handled explicitly by user or determined automatically by a Voice Activity Detection (VAD)~\cite{pywebrtcvad} module, which detects when the user has finished speaking and signals the system to begin its response—or vice versa. Traditionally, spoken dialogue systems~\cite{glass1999challenges,young2013pomdp} comprised a cascaded pipeline of multiple modules.
Motivated by the success of large language models (LLMs)~\cite{brown2020language,gpt4}, recent works have introduced speech+text LMs (SLMs)~\cite{nguyen2024spirit,jinchuan2024speechlm,arora2025landscapespokenlanguagemodels} that model the
joint distribution of speech and corresponding text, typically trained using paired
(speech, text transcription) data. 
These SLMs can be used to construct end-to-end (E2E) SDS by directly generating spoken responses from speech input in a single architecture, thus avoiding error propagation and better capturing non-phonemic information.
\begin{figure*}[t]
\centering
    \begin{subfigure}{0.32\textwidth}
        \centering
        \includegraphics[width=\linewidth]{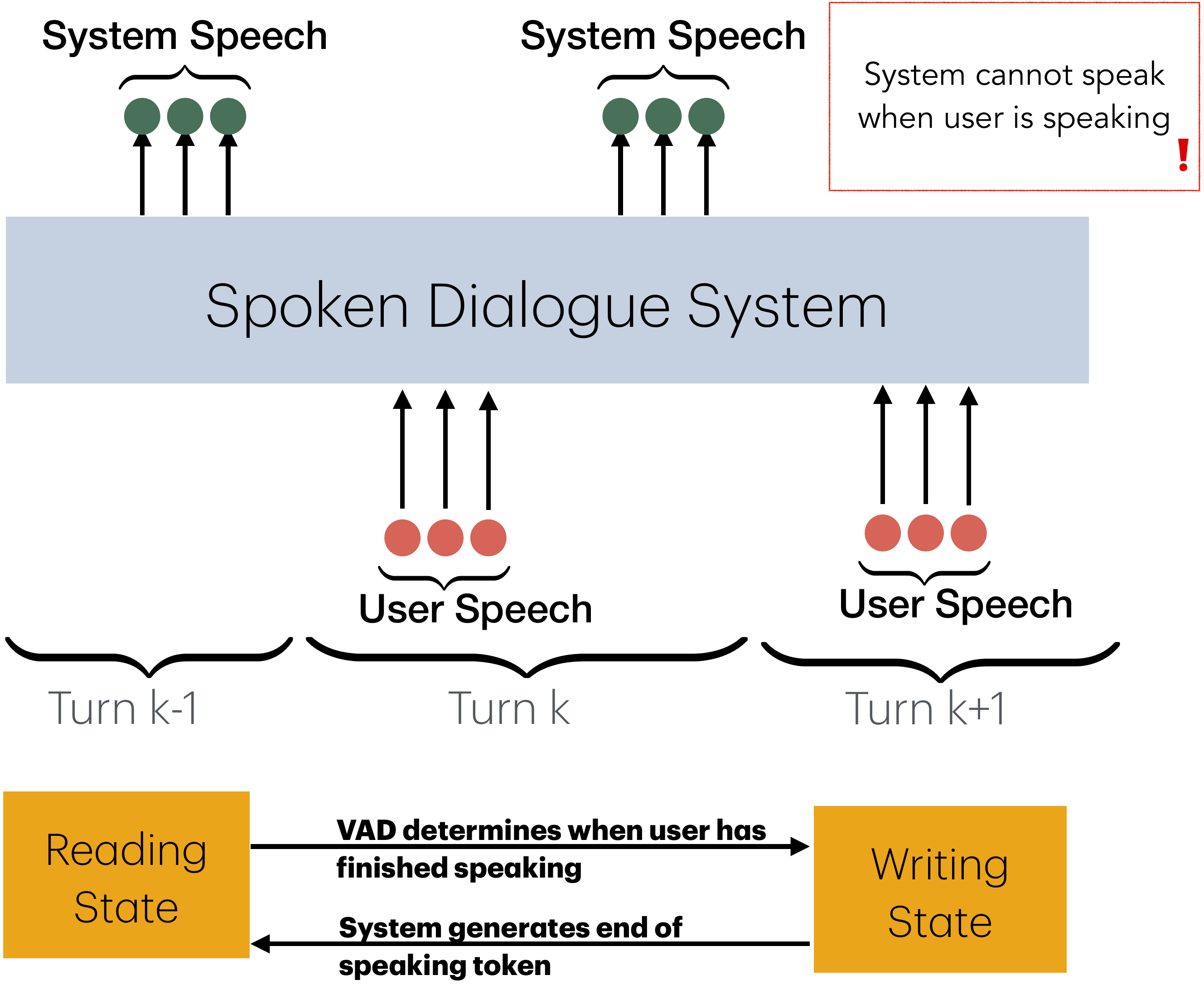} %
        \caption{Turn-by-turn SDS}
        \label{fig:turn-by-turn-approach}
    \end{subfigure}
    \begin{subfigure}{0.32\textwidth}
        \centering
        \includegraphics[width=\linewidth]{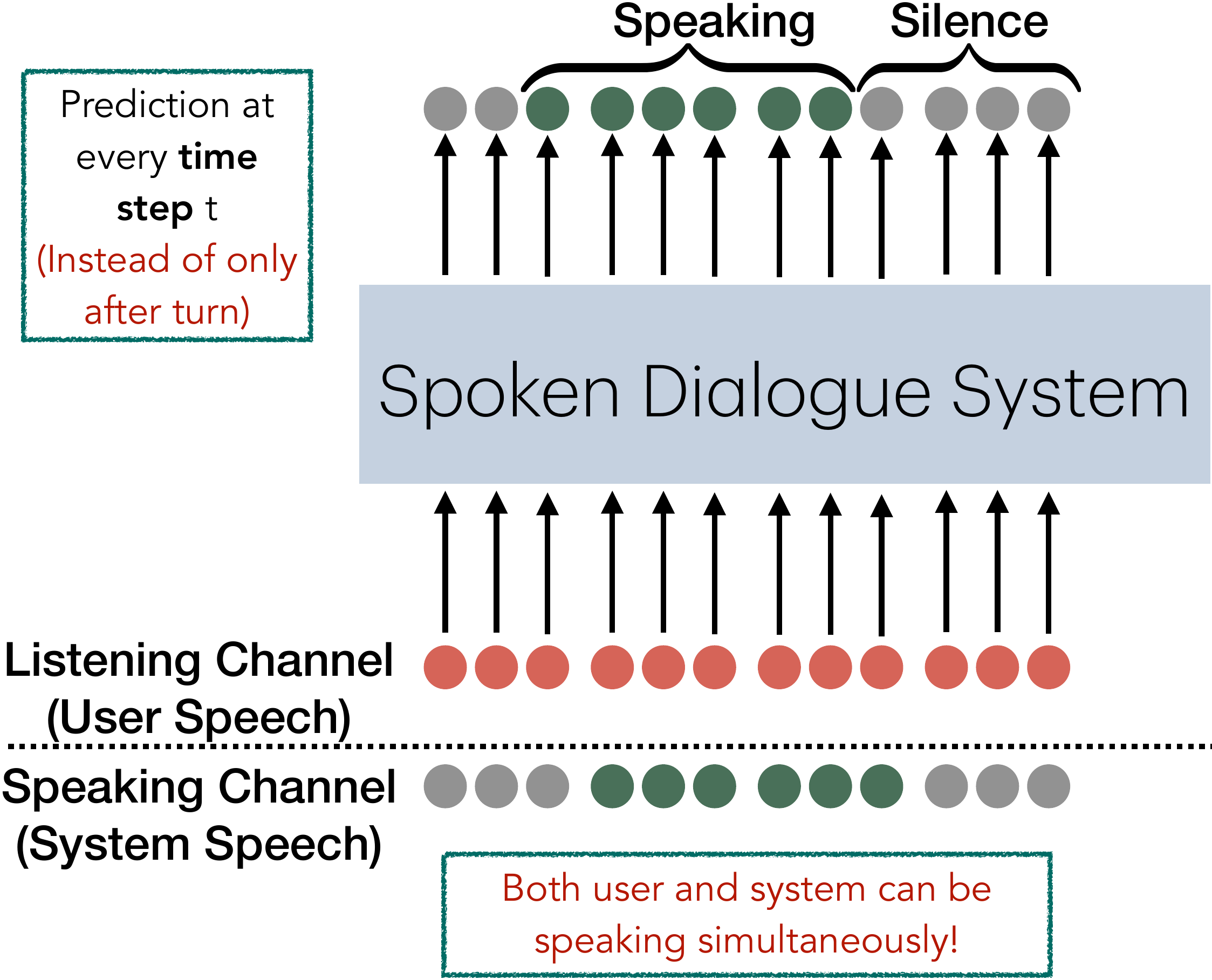} %
        \caption{Duplex Dual Channel SDS}
        \label{fig:duplex-dual-channel-approach}
    \end{subfigure}
    \begin{subfigure}{0.32\textwidth}
        \centering
        \includegraphics[width=\linewidth]{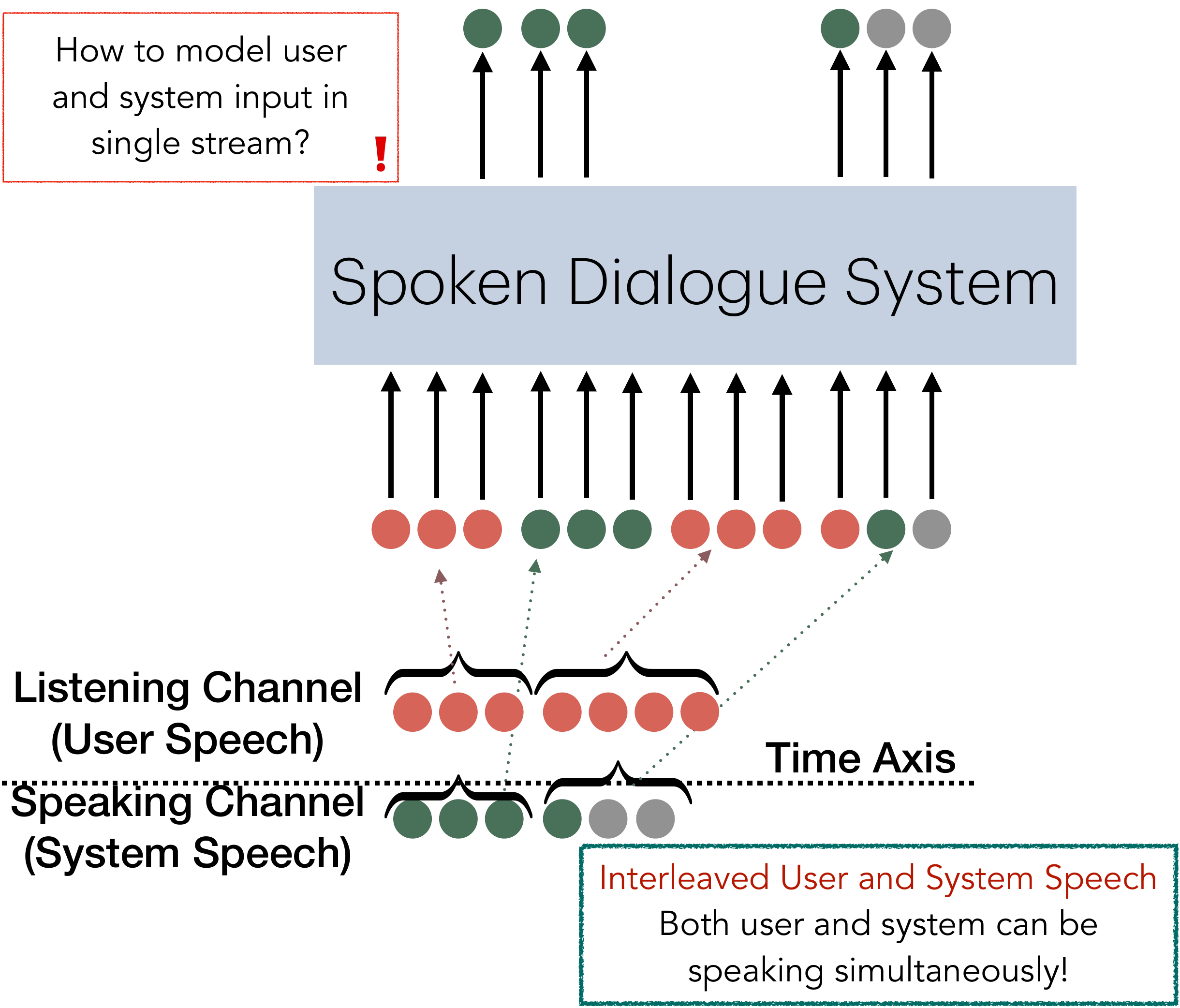} %
        \caption{Duplex Time Multiplexing SDS}
        \label{fig:duplex-single-channel-approach}
    \end{subfigure}
\caption{ Comparison of spoken dialogue system (SDS) paradigms.
(a) Turn-by-turn SDS: user and system alternate with strict turn boundaries.
(b) Duplex Dual Channel SDS: separate listening and speaking channels allow simultaneous speech generation.
(c) Duplex Time Multiplexing SDS: user and system speech are interleaved in a shared stream. }
\vskip -0.1in
\label{fig:system-overview_intro}
\end{figure*}

However, most turn-by-turn SDS architectures~\cite{huggingface_speech_to_speech,xie2024miniomnilanguagemodelshear} still rely on an external VAD module to detect silence regions and trigger turn-taking. This is problematic because silence alone is not the main cue for humans to switch turns. \cite{ten2005temporal} shows that silences within a speaker’s own utterances are often longer than those between speakers, making it difficult for VAD to distinguish between a pause and an actual turn completion~\cite{arora2025talkingturnsbenchmarkingaudio}. Moreover, spoken dialogue is inherently bidirectional: users expect the system to listen and speak simultaneously, as in natural human-human conversations~\cite{TRP1,ma2024languagemodellistenspeaking}. In contrast, VAD-driven turn-by-turn systems operate in a single mode—either listening or speaking—making them incapable of generating overlapping speech.

Inspired by this, prior works~\cite{Dialog_GSLM,meng2024parrot,zhang2024turnbasedgameenablingrealtime
} have proposed duplex architectures that eliminate the need for explicit turn-taking mechanisms, such as
VAD modules and instead predicts the output at every timestep, including the silence tokens. However, these duplex E2E models often require a large amount of training data (Moshi~\cite{kyutai2024moshi} is trained using 7 million hours of unlabelled audio data for multi-stream post-training and 20K hours of speech conversation data.).   

Recent studies~\cite{arora2025cotsds,zhang2023speechgpt,zhang2024omniflatten} have explored the application of the chain-of-thought (CoT) training paradigm~\cite{wei2022chain,zhang2022automatic} to develop turn-by-turn E2E SDS. These works demonstrate that CoT-based formulations can improve training efficiency and generate more semantically coherent and interpretable responses compared to standard E2E systems. Additionally, they offer greater parameter efficiency and enhanced ability to capture non-phonemic information relative to traditional cascaded architectures.
Motivated by these findings, we extend this paradigm to the duplex setting, proposing \textbf{\emph{SCoT}, a Streaming CoT framework for duplex E2E SDS}. The key contributions of this work are as follows:
\begin{itemize}
    \item We propose \emph{SCoT}, a CoT training framework for streaming duplex E2E spoken dialogue systems, enabling simultaneous listening and speaking without VAD-based turn segmentation. To our knowledge, we are the \emph{first} to adapt CoT-style structured reasoning, with intermediate ASR and text response targets, into a blockwise streaming duplex SDS framework.
    \item  We introduce a novel use of CTC-based forced alignments to generate block-level intermediate targets for both ASR and text response generation. This alignment-driven formulation enables tight coupling of speech and text reasoning in a way not explored by OMNI-Flatten~\cite{zhang2024omniflatten} or other prior duplex models.
    \item  Our results show that \emph{SCoT} produces more intelligent, coherent, and contextually appropriate responses, while also delivering greater training efficiency and more faithfully preserving speaking styles compared to traditional duplex SDS approaches.
    \item Further, SCoT achieves lower-latency interactions and can generate overlapping speech, outperforming prior turn-by-turn CoT-based SDS in terms of conversational fluidity and responsiveness.
    \item Finally, we will open-source our code and trained models to facilitate future development on training full duplex E2E dialogue systems. 
\end{itemize}

\section{Problem Formulation}
\label{sec:methods}
SDS take a $d$-dimensional continuous feature stream $X = (\mathbf{x}_t \in \mathbb{R}^d \mid t = 1, \dots, T)$ as input audio from the user and generate a synchronized spoken response $Y^{\text{sds}} = (\mathbf{y}^{\text{sds}}_t \in \mathbb{R}^d \mid t = 1, \dots, T)$, where $T$ is the total duration of the conversation, including multiple turns. These systems aim to estimate the posterior $P(Y^{\text{sds}}|X, X^{\text{spk}})$ using maximum a posteriori (MAP) decision theory, where $X^{\text{spk}}$ is the speaker prompt controlling the output voice characteristics.

Further, SDS can be broadly categorized into two classes: \emph{simplex} (half-duplex) or \emph{duplex} (full-duplex) systems. In simplex systems, the system can either listen or speak, but not both simultaneously. Thus at any time step $t$, simplex SDS satisfies:
\begin{equation}
\begin{cases}
x_t \neq \varnothing, \; y^{\text{sds}}_t = \varnothing & \text{(listening mode)} \\[6pt]
x_t = \varnothing, \; y^{\text{sds}}_t \neq \varnothing & \text{(speaking mode)} 
\end{cases}
\label{eq:simplex_definition}
\end{equation}
where $\varnothing$ indicates silence. In contrast, duplex systems can listen and speak at the same time, enabling overlapping speech as shown below:
\begin{equation}
x_t \neq \varnothing, \quad y^{\text{sds}}_t \neq \varnothing
\label{eq:duplex_definition}
\end{equation}

\section{Turn-by-turn (Simplex) SDS}
\label{subsec:turn-by-turn}

\begin{figure*}[t]
\centering
    \includegraphics[width=\linewidth]{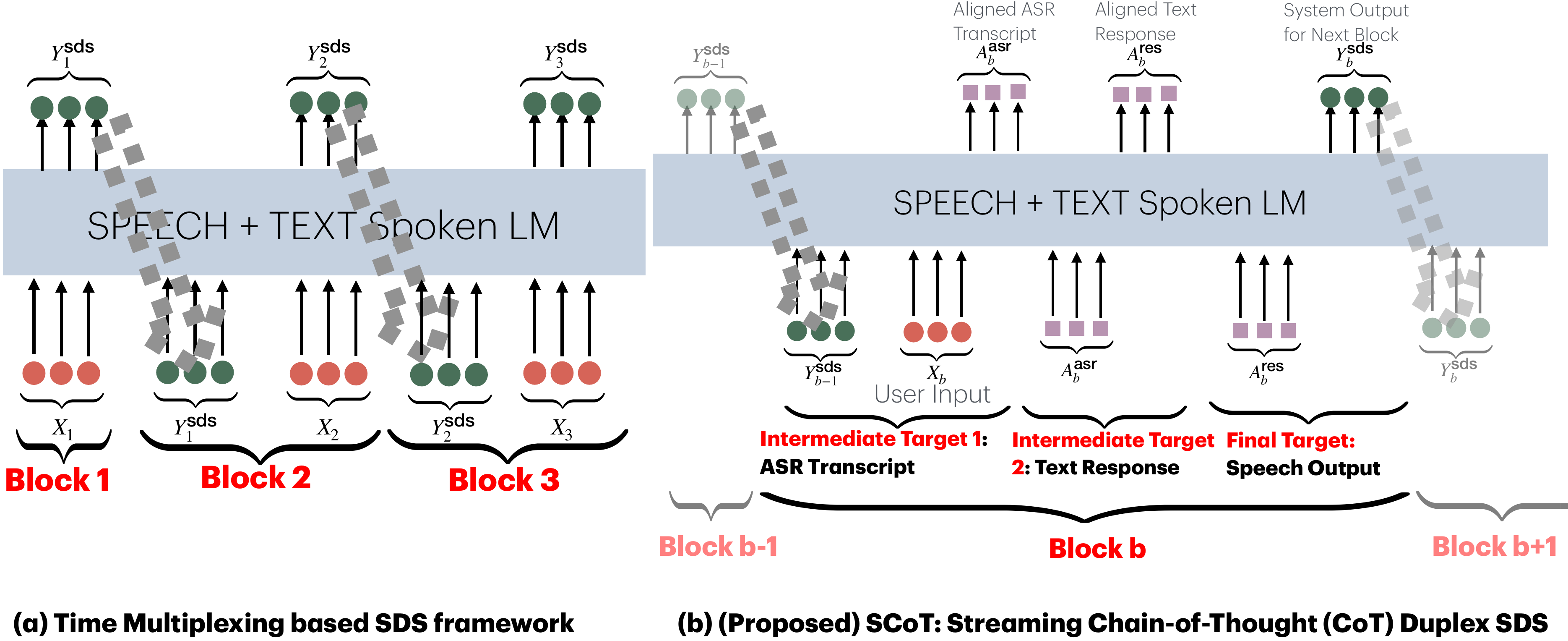} %
\caption{Overview of \emph{SCoT}: the proposed Streaming Chain-of-Thought (CoT) Duplex SDS architecture (Sec.~\ref{AA}). Building on the time-multiplexing framework (Sec.~\ref{subsec:duplex_SDS}), SCoT alternates between processing fixed-duration segments of user input and generating speech output, enabling single-channel, real-time duplex interaction. At each block, the model additionally predicts aligned intermediate textual representations—namely, the ASR transcript and text response—prior to generating the final speech output, improving semantic coherence of output and training efficiency.}
\vskip -0.2in
\label{fig:proposed-cot-multiplexing-approach}
\end{figure*}
SDS~\cite{glass1999challenges} have traditionally 
operated in a turn-by-turn manner as shown in Fig.~\ref{fig:turn-by-turn-approach}.
This strict turn-based setting corresponds to a \emph{simplex} interaction pattern 
(Eq.~\ref{eq:simplex_definition}), 
where at any given time the system is restricted to either \emph{listening mode} or \emph{speaking mode}, 
but never both simultaneously. 
Turns are obtained by segmenting a $T$-length conversation with a VAD module, which identifies speech boundaries to determine whether the system should listen or respond. 
Building on the global problem formulation in Sec.~\ref{sec:methods}, the continuous input stream $X$ and corresponding output $Y^{\text{sds}}$ are segmented into variable-length turns i.e. $Y^{\text{sds}}=\{Y^{\text{sds}}_k \mid k=1, \dots, K\}$ and $X=\{X_k \mid k=1, \dots, K\}$ where $K$ denotes the total number of turns. Note that the duration of each turn is highly variable, depending on the speaker’s behavior and the conversation context. 

Thus, the user’s $k$-th turn is defined as any speech activity beginning after $T_{k-1}$ and ending at $T_{k}$, where $T_{k}$ is the first point at which at least $\delta$ consecutive silence frames are detected by VAD.
Here, $\delta$ is a hyperparameter.
The output of the VAD module is then used to create a \emph{buffered speech utterance} $X_k$ containing user input for the current turn:
\begin{equation}
X_k = \mathbf{x}_{T_{k-1} : T_{k} - \delta}
\label{eq:intro_sds_vad}
\end{equation}
The SDS system then outputs the speech response $Y^{\text{sds}}_{k}= (\mathbf{y}^{\text{sds}}_t \in \mathbb{R}^d \mid t = T_{k}, \dots, T_{k}+T')$ where $T'$ is length of speech response at the current turn. After completing the response, the system re-enters listening mode at time $T_{k}+T'$.

Given this turn-based interaction, the system models the overall speech output distribution as:
\begin{equation}
    P(Y^{\text{sds}}|X, X^{\text{spk}})= \prod_{k} P(Y^{\text{sds}}_k|Y^{\text{sds}}_{1:k-1},X,X^{\text{spk}}) \label{eq:intro_sds_turn}
\end{equation}
Introducing conditional independence (C.I.) assumption, where each response depends only on the current and past inputs causally, we obtain:
\begin{equation}
    P(Y^{\text{sds}}|X, X^{\text{spk}})= \prod_{k} P(Y^{\text{sds}}_k|Y^{\text{sds}}_{1:k-1},X_{1:k},X^{\text{spk}}) \label{eq:intro_sds_turn_CI}
\end{equation}
Thus, the dialogue distribution is expressed as a product of turn-level conditional probabilities, and the objective of a turn-by-turn SDS is to predict, at each turn, the response that maximizes this probability. These turn-level conditional probabilities can be optimized in multiple ways, as discussed in the following subsections.

\subsection{Turn-by-turn Cascaded SDS}
Traditionally, spoken dialogue systems~\cite{glass1999challenges,huang2024audiogpt} realized the turn-level conditional probability in Eq.~\ref{eq:intro_sds_turn_CI} using a cascaded pipeline. This pipeline typically comprised multiple modules, including automatic speech recognition (ASR), natural language understanding~(NLU), natural language generation (NLG), and text-to-speech (TTS) synthesis. However, since each module in this cascaded pipeline is separately optimized, the system suffers from error
propagation and often converges to suboptimal overall performance.

\subsection{Turn-by-turn E2E SDS}
Recent efforts have focused on reducing the number of modules required to build conversational AI systems in an end-to-end manner. 
Prior works~\cite{zhang2023speechgpt,zhang2024speechgpt,nguyen2024spirit} adopt SLMs to directly estimate the output speech $\hat{Y}^{\text{sds}}$ within a single autoregressive architecture by modeling $P(Y^{\text{sds}}_k|Y^{\text{sds}}_{1:k-1},X_{1:k},X^{\text{spk}})$, as outlined in Eq.~\eqref{eq:intro_sds_turn_CI}. A common strategy for fine-tuning such models involves synthesizing conversational speech data using a TTS system applied to text-based dialogue corpora~\cite{zhang2023speechgpt}. While these systems avoid the error propagation typical of cascaded systems, they often lack structured reasoning~\cite{arora2025cotsds}, which can lead to less coherent responses and significantly higher training data requirements~\cite{kyutai2024moshi}.

\subsection{Turn-by-turn CoT E2E SDS}
Recent efforts have looked into addressing these limitations through Chain-of-Thought formulation by incorporating ASR transcript $S^\text{asr}_{k}$ and text response $S^\text{res}_{k}$ at each turn within the E2E spoken dialogue formulation via the sum rule. Using the Viterbi approximation and C.I. assumption, we can modify Eq.~\eqref{eq:intro_sds_turn_CI} to get:
\begin{equation}
P(Y^{\text{sds}}|X, X^{\text{spk}})\approx \prod_{k} P(Y^{\text{sds}}_k|Y^{\text{sds}}_{1:k-1},X_{1:k},\hat{S}^{\text{res}}_{1:k}, \hat{S}^{\text{asr}}_{1:k}, X^{\text{spk}}) \label{eq:turn_cot_final_formulation}
\end{equation}
where
\begin{equation}
\hat{S}^{\text{asr}}_{k} = \argmax_{S^{\text{asr}}_{k}} p(S^{\text{asr}}_{k}|Y^{\text{sds}}_{1:k-1},X_{1:k},\hat{S}^{\text{res}}_{1:k-1}, \hat{S}^{\text{asr}}_{1:k-1}, \cancel{X^{\text{spk}}})\label{eq:turn_cot_asr_formulation}
\end{equation}
and
\begin{equation}
\hat{S}^{\text{res}}_{k} = \argmax_{S^{\text{res}}_{k}} p(S^{\text{res}}_{k}|Y^{\text{sds}}_{1:k-1},X_{1:k},\hat{S}^{\text{res}}_{1:k-1}, \hat{S}^{\text{asr}}_{1:k}, \cancel{X^{\text{spk}}})\label{eq:turn_cot_t2t_formulation}
\end{equation}
At each turn, the \textbf{CoT model} follows a structured decoding process: \circled{1} first generating the ASR transcript $\hat{S}^{\text{asr}}_{k}$ using Eq.~\eqref{eq:turn_cot_asr_formulation}, \circled{2} then
predicts the text response $\hat{S}^{\text{res}}_{k}$ using Eq.~\eqref{eq:turn_cot_t2t_formulation} and \circled{3} generates the final speech output $\hat{Y}^{\text{sds}}_k$ using these intermediate outputs (Eq.~\eqref{eq:turn_cot_final_formulation}). This CoT E2E model can be implemented similarly using a speech+text language model that performs each stage of autoregressive decoding sequentially according to Eqs.~\eqref{eq:turn_cot_final_formulation}--\eqref{eq:turn_cot_t2t_formulation}.

Despite this progress, these systems still depend on an external VAD\footnote{Other approaches~\cite{castillo-lopez-etal-2025-survey,ekstedt2022voiceactivityprojectionselfsupervised} have also been proposed, though they may introduce added complexity or latency.} (Eq.~\eqref{eq:intro_sds_vad}) to perform turn-taking, i.e., knowing when to speak up. However, this is insufficient for having \emph{natural} conversations because silence is often not the main cue for humans to switch turns~\cite{ten2005temporal}, as we discussed in Sec.~\ref{sec:intro},  and further cannot generate overlapping speech common in human-human interactions. 

\section{Duplex system}
\label{subsec:duplex_SDS}

There have been efforts to build \emph{duplex} or \emph{speaking while listening} spoken dialogue systems~\cite{EOT1,turntakeingIS18,skantze-2017-towards,MEENA2014903,EOT2}. A duplex spoken dialogue system (SDS) is a system that can speak and listen simultaneously (Eq.~\ref{eq:duplex_definition}), enabling real-time, natural conversations with users—much like how humans interact in daily life. 
Similar to other E2E SDS, these duplex SDS leverage pre-trained SLMs as their backbone, enabling autoregressive generation across both speech and text modalities.

\subsection{Dual Channel}
\label{subsec:dual_channel}
One way to achieve duplex dialogue is by using dual channels~\cite{Dialog_GSLM,kyutai2024moshi,ma2024languagemodellistenspeaking,meng2024parrot} as shown in Fig.~\ref{fig:duplex-dual-channel-approach}. 
The SLM has two input channels: the listening channel which continuously receives input, and the speaking channel where the spoken output from the SLM is directed, allowing the model to track what it has said. 
Unlike simplex systems that operate only at discrete turn boundaries (Eq.~\eqref{eq:intro_sds_turn_CI}), this approach generates responses continuously at every timestep 
$t$ (Sec.~\ref{sec:methods}), eliminating the need for explicit turn-taking mechanisms (Eq.~\eqref{eq:intro_sds_vad}): 
\begin{equation}
P(Y^{\text{sds}}|X, X^{\text{spk}})= \prod_{t}~P(\mathbf{y}^{\text{sds}}_{t} |\mathbf{y}^{\text{sds}}_{1:(t-1)},\mathbf{x}_{1:t},\cancel{\mathbf{x}_{t+1:T}}, X^{\text{spk}}).\label{eq:intro_sds_duplex}
\end{equation}
This formulation corresponds to a continuous, frame-level duplex system, 
since at every time step $t$, both $x_t$ and $y^{\text{sds}}_t$ may not be silent
(i.e., $x_t \neq \varnothing, \; y^{\text{sds}}_t \neq \varnothing$ similar to Eq.~\ref{eq:duplex_definition}).
However, the dual channel formulation significantly diverges from SpeechLM's pre-training setup and therefore demands extensive training compute.

\subsection{Time Multiplexing}
An alternative line of work~\cite{wang2024freezeomnismartlowlatency,xu2024enablingrealtimeconversationsminimal,duplextimeNuerIPS24} explores a single-channel approach, where the SLM operates with a single channel and must explicitly alternate between listening and speaking modes as shown in Fig.~\ref{fig:duplex-single-channel-approach}. A key advantage of this setup is that it is better aligned with the SLM's pre-training setup~\cite{jinchuan2024speechlm}. The primary challenge, however, lies in determining when to switch between listening and speaking. 

One common strategy is the time-multiplexing approach, which alternates between processing \emph{fixed}-duration segments of user input—unlike the variable-length speaker turns used in Section~\ref{subsec:turn-by-turn}—and generating spoken output for the subsequent segment~\cite{veluri2024beyond}, as illustrated in Fig.~\ref{fig:proposed-cot-multiplexing-approach}(a).
 In this setup,  the input speech X is divided into a sequence of $B$ blocks, $X = \{X_b \mid b=1, \dots, B\}$. Assuming a block size of $N_{\text{block}}$,  the $b^{th}$ block is $X_{b}=\mathbf{x}_{(I_{b-1}+1):I_{b}}$ where $I_{b-1}=(b-1)*N_{\text{block}}$. 
Similarly, the output speech $Y^{\text{sds}}$ is represented as a sequence of $B$ blocks, $Y^{\text{sds}} = \{Y^{\text{sds}}_b \mid b=1, \dots, B\}$.
Rather than producing output at every frame $t$ as in Eq.~\eqref{eq:intro_sds_duplex}, the model generates output block-by-block in a streaming fashion, estimating the posterior using the causality-based C.I. as:
\begin{equation}
P(Y^{\text{sds}}|X, X^{\text{spk}})= \prod_{b}~P(Y^{\text{sds}}_b|Y^{\text{sds}}_{1:b-1},X_{1:b},\cancel{X_{b+1:B}},X^{\text{spk}}).\label{eq:intro_sds_time_multiplex}
\end{equation}
This formulation corresponds to a blockwise duplex system, 
where both $X_b$ and $Y^{\text{sds}}_b$ may be non-silent within the same block. 
Thus, although speech generation is chunked into blocks rather than frame-by-frame (Sec. IV-\ref{subsec:dual_channel}), 
the system still supports simultaneous listening and speaking, 
consistent with the definition of duplex (Eq.~\ref{eq:duplex_definition}), while retaining the architectural simplicity of a single-channel, autoregressive design.
Hence, it enables incremental speech generation, providing a balance between responsiveness and coherence by remaining consistent with SpeechLM’s pre-training setup.

\section{Our Chain-of-Thought Duplex SDS}
\label{AA}
Building on the insights from the turn-by-turn CoT SDS framework (Eq.~\eqref{eq:turn_cot_final_formulation}-\eqref{eq:turn_cot_t2t_formulation}), we propose \emph{SCoT}: a fully duplex SDS architecture that incorporates the CoT formulation within the time-multiplexing formulation (Eq.~\eqref{eq:intro_sds_time_multiplex}) as shown in Fig.~\ref{fig:proposed-cot-multiplexing-approach}(b). 
In this blockwise streaming CoT setting, we can no longer rely on the full ASR transcript and full-text response as intermediate representations, as in Eq.~\eqref{eq:turn_cot_final_formulation}. Instead, we require \emph{fine-grained alignment} between these textual elements and their associated speech segments to infer the transcript and system response within each block.
Let $A^{\text{asr}} = \{a^{\text{asr}}_t \mid t=1, \dots, T\}$ be the frame-level alignment\footnote{These frame-level alignments may include repeated and blank tokens.} of the ASR transcript $S^{\text{asr}}$, where each $a^{\text{asr}}_t$ denotes the aligned transcript token for speech frame $\mathbf{x}_t$. Similarly, let $A^{\text{res}} = \{a_t^{\text{res}} \mid t=1, \dots, T\}$ represent the frame-level alignment of the generated text response $S^\text{res}$. 
Let the corresponding aligned transcript and system response for the $b^{\text{th}}$ block (See Sec.~\ref{subsec:duplex_SDS}) be $A^{\text{asr}}_b=a^{\text{asr}}_{I_{b-1}+1:I_{b}}$ and $A^{\text{res}}_b=a^{\text{res}}_{I_{b-1}+1:I_{b}}$ respectively.
Incorporating these alignments, we extend the CoT formulation (Eqs.~\eqref{eq:turn_cot_final_formulation}-\eqref{eq:turn_cot_t2t_formulation} in Sec.~\ref{subsec:turn-by-turn}) into the duplex setting as follows:
\begin{align}
P(Y^{\text{sds}}|X, X^{\text{spk}}) &\approx \prod_{b} P(Y^{\text{sds}}_b|Y^{\text{sds}}_{1:b-1},X_{1:b}, \hat{A}^{\text{asr}}_{1:b}, \hat{A}^{\text{res}}_{1:b},X^{\text{spk}}).\label{12_duplex_sds_eq:cot}
\end{align}
The intermediate text response is predicted as:
\begin{equation}
    \hat{A}^{\text{res}}_{b} = \argmax P({A}^{\text{res}}_{b}|Y^{\text{sds}}_{1:b-1},X_{1:b},\hat{A}^{\text{asr}}_{1:b}, \hat{A}^{\text{res}}_{1:b-1})\label{12_duplex_sds_eq:t2t}
\end{equation}
and the aligned ASR transcript for each block is given by:
\begin{equation}
\hat{A}^{\text{asr}}_{b} = \argmax P({A}^{\text{asr}}_{b}|Y^{\text{sds}}_{1:b-1},X_{1:b},\hat{A}^{\text{asr}}_{1:b-1}, \hat{A}^{\text{res}}_{1:b-1}).\label{12_duplex_sds_eq:asr}
\end{equation}
At each block, SCoT performs a structured three-stage decoding process, similar to Sec.~\ref{subsec:turn-by-turn}. During training, we utilize CTC-based forced alignment to generate the intermediate targets ${A}^{\text{res}}_{b}$ and ${A}^{\text{asr}}_{b}$. 

Since the optimal CoT reasoning path is unknown, we introduce three \emph{SCoT} variants: (i) \emph{SCoT-ASR}, where only the ASR transcript (i.e., $\hat{A}^{\text{asr}}_{b}$ in Eq.~\ref{12_duplex_sds_eq:asr}) is used as the intermediate target; (ii) \emph{SCoT-Response}, where only the text response (i.e., $\hat{A}^{\text{res}}_{b}$ in Eq.~\ref{12_duplex_sds_eq:t2t}) serves as the intermediate target; and (iii) \emph{SCoT-Full}, where both the transcript and text response are jointly incorporated as intermediate targets, as in Eq.~\ref{12_duplex_sds_eq:cot}. This design enables the \emph{first} systematic investigation into the relative utility of different CoT reasoning paths in a duplex SDS setting, highlighting their trade-offs between semantic coherence and interaction latency.

Compared to the standard turn-by-turn CoT formulation (Eq.~\eqref{eq:turn_cot_final_formulation}), our blockwise streaming duplex approach enables significantly lower latency, thus supporting real-time interaction, and is capable of handling overlapping speech. In contrast to conventional time-multiplexed duplex systems (Eq.~\eqref{eq:intro_sds_time_multiplex}), our method incorporates structured reasoning, which improves semantic coherence, as demonstrated in Sec.~\ref{sec:results}. Unlike OMNI-Flatten~\cite{zhang2024omniflatten}, \emph{SCoT} uses CTC forced alignments to explicitly map intermediate text targets with speech response. This strategy aligns each reasoning stage with the SpeechLM’s pre-training tasks—namely ASR, text LM, and TTS — leading to faster convergence and improved training data efficiency~\cite{arora2025cotsds}.

\section{Experiments}

\subsection{Datasets}
For our experiments, we focus on \emph{real} human-human conversation datasets, selecting 2 widely used corpora: Switchboard~\cite{Switchboard} ($\approx$ 300 hours) and Fisher~\cite{cieri2004fisher} ($\approx$ 2000 hours). 
For Switchboard, we use the Eval2000 dataset for evaluation. We train our baseline and duplex models in a \emph{combined} setting (SWBD+Fisher train) using the entire Fisher dataset and the Switchboard training set and evaluate it on the Eval2000 dataset similar to prior work~\cite{arora2025cotsds}.

We evaluate spoken dialogue systems across semantic quality and audio quality. For semantic quality, we report ROUGE~\cite{lin2004rouge} and METEOR~\cite{banerjee2005meteor} scores, using human references as ground truth, alongside perplexity~\cite{jelinek1977perplexity}, computed using GPT-2~\cite{gpt-2}.
We additionally evaluated our system using Qwen2.5-7B-Omni as an LLM-based judge, following the protocol in recent work~\cite{zhang2024omniflatten}. 
For E2E models, semantic quality is evaluated by transcribing synthesized speech $\hat{Y}$ using Whisper large ~\cite{whisper}. 

We utilize the VERSA toolkit~\cite{shi2024versa}, measuring audio quality using UTMOS~\cite{saeki22c_interspeech}. 
To demonstrate our system's ability to handle overlapping speech, we report the percentage of blocks where overlapping output is generated. By comparing with overlapping speech in reference human-human dialogues, we also compute precision and recall, indicating how accurately the system reproduces natural overlaps.
We compute latency by performing inference using single H100 GPU.
We follow the setup in~\cite{ji2024wavchatsurveyspokendialogue} and measure the time taken to generate the first token of speech (i.e., the waiting time after the user finishes speaking) and compute the overall Real-Time Factor (RTF) by dividing it with the total duration of the input speech segment. RTF values are computed over a representative subset of 300 turn-level utterances from the test set across all models.
We evaluate speaking-style consistencies within the entire conversation by extracting emotion vectors from both synthesized outputs and ground-truth responses using Emo2Vec~\cite{ma2023emotion2vec}, then measure their cosine similarity. 
We rank all spoken dialogue systems based on their emotional alignment and compute the average rank (``Emotion Rank''~\cite{arora2025cotsds}) across all utterances. 
\begin{table*}[t]
    \centering
    \caption{Semantic quality evaluation. The performance of SmolLM is generated using ground-truth transcripts. SCoT significantly improves semantic quality over standard time-multiplexing Duplex E2E model. SCoT significantly reduces latency comparable to turn-by-turn CoT E2E models.\xmark : Prior work~\cite{arora2025cotsds} does not report results with Qwen-OMNI as a judge as well as RTF. We compute latency for turn-by-turn E2E model for comparison.}
    {
    \begin{tabular}{l|ccccc}
        \hline
        \textbf{Model} & \multicolumn{3}{c}{SWBD} \\
        & \textbf{ROUGE-1/2/L (↑)} & \textbf{METEOR (↑)} & \textbf{Perplexity (↓)} & \textbf{Qwen-OMNI judge (↑)} & \textbf{RTF (↓)}\\
        \hline
        SmolLM v2-1.7B (GT)~\cite{arora2025cotsds} & 14.0 / 2.3 / 11.5 & 12.9 & \hphantom{0}25.4 &  \xmark &  \xmark\\
        E2E~\cite{arora2025cotsds} & 10.5 / 0.8 / \hphantom{0}8.4 & \hphantom{0}9.7 & 302.2 &  \xmark & 0.75 \\\midrule
        CoT E2E~\cite{arora2025cotsds} & 14.2 / 1.3 / 10.3 & 12.9 & \hphantom{0}28.5 &  \xmark &  \xmark\\
        \hphantom{0}SWBD+Fisher train~\cite{arora2025cotsds} & 14.6 / 1.5 / 10.4 & 14.9 & \hphantom{0}22.2 &  \xmark &  \xmark\\
        \hphantom{00}Multi turn & 15.4 / 1.7 / 12.1 & 14.2 & \hphantom{0}\textbf{20.7} & \textbf{4.02} & 2.95\\\midrule
        Duplex E2E & 14.4 / 1.4 / 10.3 & 11.8 & 313.8 & 2.20 & 0.29 \\ \midrule
        \emph{SCoT-ASR} (proposed)  &  11.1 / 1.1 / \hphantom{0}8.1 & \hphantom{0}8.5 & 432.7 & 2.19& 0.56\\
        \emph{SCoT-Response} (proposed)  & \textbf{27.9} / \textbf{6.8} / \textbf{19.8} & \textbf{26.4} & \hphantom{0}42.3 & 3.24 & 0.56\\
        \emph{SCoT-Full} (proposed)  & 22.3 / 4.2 / 15.8 & 20.3  & \hphantom{0}55.2 & 2.76& 0.85 \\
        \hline
    \end{tabular}
    }
    \label{tab:text_response_results}
\end{table*}

\subsection{Baseline}
Similar to ~\cite{arora2025cotsds}, we report performance of strong task-specific baselines\footnote{Baselines are not fine-tuned on spoken dialogue datasets.}: SmolLM v2 1.7B-Instruct~\cite{allal2025smollm2smolgoesbig} for text generation and the single-speaker TTS model (\textbf{LJSpeech VITS})~\cite{hayashi2020espnet} from ESPnet-TTS. For audio quality, we evaluate pre-trained SLM in two settings: zero-shot and fine-tuned.

We additionally report the performance of single turn cascaded and E2E systems that were developed in~\cite{arora2025cotsds}. The cascaded systems combine: SpeechLM (ASR), SmolLM v2 (text response), and SpeechLM (TTS), evaluated in both zero-shot (\textbf{Cascaded (Pre-train)}) and fine-tuned (\textbf{Cascaded (Fine-tune)}) modes.
We also compare with a traditional E2E system (\textbf{SpeechLM E2E} in \cite{arora2025cotsds}), where SpeechLM is trained end-to-end to directly predict speech outputs as well as \emph{Chain-of-Thought} single turn E2E SDS~\cite{arora2025cotsds}.

We first extend the CoT SDS setup to a multi-turn setting using Eq.~\eqref{eq:turn_cot_final_formulation}-\eqref{eq:turn_cot_t2t_formulation} in Sec.~\ref{subsec:duplex_SDS} to establish a strong baseline for our work. Each training instance includes up to four dialogue turns or a maximum duration of 120 seconds, whichever is shorter. We simulate either speaker in the human-human conversation as the assistant, thus creating 2 separate instances; however, if the assistant is simulated as the first speaker, we do not predict its initial utterance.  During inference, we provide the ground-truth context from the previous $k-1$ turns and generate the model's response for the $k^{th}$ turn.

To build the duplex end-to-end SDS system, we adopt a time-multiplexing approach (Sec.~\ref{subsec:duplex_SDS},~\cite{veluri2024beyond}) with a block size $N_{\text{block}}=2$ seconds. Human-human conversation audio is segmented into 60-second dialogue contexts to construct training instances.
During inference, we follow the same setup as in the multi-turn setting: the model is provided with the ground-truth context from the previous $k-1$ blocks (corresponding to 29 previous blocks) and is tasked with predicting the system response for the next block. Finally, we train the 3 variants of \emph{SCoT}, our CoT E2E Duplex SDS by computing word-level timestamp using forced alignments from OWSM CTC v4 1B~\cite{owsm-v4}. For words that lie across block boundaries, we include them as intermediate targets in both corresponding blocks to ensure continuity and alignment consistency. Both multi-turn CoT and duplex SDS models are trained on a combined dataset comprising the Fisher and Switchboard training sets.

\subsection{Experimental Setups}
Our models are implemented in PyTorch, with all experiments conducted using the ESPnet~\cite{espnet,ESPnet-SLU,arora2025espnet} toolkit. 
We use the pre-trained SLM from~\cite{jinchuan2024speechlm} that leverages the SmolLM2 1.7B text LLM for initialization. 
We adopt the delay interleave architecture~\cite{musicgen} for multi-stream language modeling.
For audio tokenization, we utilize ESPnet-Codec~\cite{shi2024espnet}\footnote{\url{https://huggingface.co/ftshijt/espnet_codec_dac_large_v1.4_360epoch}} for codec tokenization and XEUS\footnote{\url{https://huggingface.co/espnet/xeus}, K-means tokenizer trained on the last-layer representation with 5k clusters}~\cite{chen2024towards} for SSL tokenization. Specifically, we concatenate codec and SSL tokens frame-by-frame. 
For decoding, ASR uses greedy search followed by post-processing (similar to \cite{arora2025cotsds}) to remove hallucinations, while text response generation employs top-k sampling (k=30, temperature = 0.8), followed by post-processing~\cite{arora2025cotsds} to remove hallucinations and constrain response length. For speech response generation in CoT models, we apply top-k sampling (same as text response), and further post-process the outputs, computing top-10 samples and selecting the speech with the highest intelligibility to generated text response $\hat{S}^{\text{res}}$ (Eq.~\eqref{eq:turn_cot_t2t_formulation}) in turn-by-turn system and $\hat{A}^{\text{res}}_{b}$ (Eq.~\ref{12_duplex_sds_eq:t2t})  in duplex systems.
During inference in duplex systems, we additionally constrain the ASR transcript and text response to a maximum of 25 words, and limit the speech output to the block duration (i.e., 2 seconds). All models are trained using 4 NVIDIA H200 GPUs.
We will publicly release data processing, training and inference details.

\begin{table}[t]
    \centering
    \caption{Turn-taking behavior analysis. We report the percentage of overlapping blocks and overlap precision/recall.}
    {
    \begin{tabular}{lccc}
        \hline
        \textbf{Model} & \textbf{\%Overlap}  & \multicolumn{2}{c}{\textbf{\%Overlap Acc.}}  \\
        & & Prec & Rec  \\
        \hline
        Cascaded (Pre-train) & 0.0 & \xmark & \xmark \\
        Cascaded (Fine-tune) & 0.0 & \xmark & \xmark \\\midrule
        E2E & 0.0  & \xmark  & \xmark \\
        Multi turn CoT E2E & 0.0  & \xmark & \xmark \\\midrule
        Duplex E2E & 46.8 & \textbf{77.7} & 64.9 \\ \midrule
        \emph{SCoT-ASR} (proposed) & 41.1 & 74.1 & 54.3 \\
        \emph{SCoT-Response} (proposed) & 63.0 & 59.5 & 66.8 \\
       \emph{SCoT-Full} (proposed) & 64.5 & 62.2 & \textbf{71.5} \\
        \hline
    \end{tabular}
    }
    \label{tab:turn_taking_results}
    
\end{table}
\section{Results and Discussion}
\label{sec:results}
\subsection{Semantic Quality}
Table~\ref{tab:text_response_results} presents the evaluation of various spoken dialogue systems using ROUGE, METEOR, and perplexity metrics. The baseline SpeechLM E2E model performs poorly as reported in \cite{arora2025cotsds}, with a high perplexity of 302.2, indicating difficulty in generating coherent responses. The SpeechLM CoT E2E models show substantial improvements across all metrics. Extending the CoT formulation to the multi-turn setting yields the best results, achieving a perplexity of 20.7 and improved ROUGE and METEOR scores, underscoring the benefits of context-aware reasoning.
To ensure a fair comparison between turn-by-turn and duplex systems, we aggregate the speech output from each block within a turn and compute semantic quality metrics based on the combined output. The SpeechLM Duplex E2E model, while enabling real-time duplex interaction, struggles with semantic consistency, as reflected in its high perplexity of 313.8, highlighting the challenge of modeling streaming interactions effectively.

Among the \emph{SCoT} variants, we find that removing the text response path (\emph{SCoT-ASR}) results in a sharp performance drop, indicating that ASR-only reasoning is insufficient for streaming duplex dialogue. In contrast, both \emph{SCoT-Response} and \emph{SCoT-Full} deliver substantial improvements over the Duplex E2E baseline across all metrics. These results highlight the effectiveness of CoT reasoning in enhancing the semantic quality of duplex SDS models. Remarkably, SCoT achieves higher alignment with human references—measured by ROUGE and METEOR—than all turn-by-turn systems and even a state-of-the-art LLM (SmolLM v2), underscoring its ability to generate more natural, human-like responses. Moreover, the superior performance of \emph{SCoT-Response} over \emph{SCoT-Full} demonstrates that text-based reasoning is the key driver of semantic coherence in duplex CoT systems. This result shows that the ASR transcript need not be explicitly predicted as an intermediate target in the CoT path, reinforcing the efficiency and strength of end-to-end CoT modeling over traditional cascaded pipelines.

We also show that the judge LLM prefers the responses generated by our SCoT-Response model (Score = 3.24) over the baseline Duplex E2E model (Score = 2.20), suggesting that our CoT formulation improves semantic coherence.
However, the LLM judge assigns a higher score to the turn-by-turn CoT E2E model (Score = 4.02), as its responses tend to be more semantically fluent—consistent with its lower perplexity\footnote{The relatively high perplexity of SCoT reflects the presence of fillers and backchanneling typical of spoken conversations, which differ significantly from the written text used to train GPT-2. Notably, turn-by-turn systems~\cite{arora2025cotsds} filtered out short utterances such as backchannels from their training data.} (20.3 in Table~\ref{tab:text_response_results}). However this comes at the cost of impractically high latency (RTF = 2.95), which hinders real-time interaction.  
In contrast, our proposed SCoT-Response model  achieves a more favorable balance, delivering significantly enhanced semantic coherence while maintaining very low latency (RTF = 0.56), even below that of standard turn-by-turn E2E systems (0.75 RTF).

\subsection{Turn Taking Metrics}
We begin by comparing systems based on the percentage of blocks with overlapping speech in Table~\ref{tab:turn_taking_results}. While 57.5\% of blocks in the Eval2000 dataset contain overlapping speech, all turn-based systems—including both cascaded and E2E variants—fail to generate overlaps by design.

For duplex systems, we apply a simple validation-based heuristic to determine when the system chooses to remain silent. For the Duplex model, if more than half of the top-k generated responses contain silence, the system is considered silent where k is decided based on validation set. The standard Duplex E2E model generates overlaps in 46.8\% of blocks, with strong precision but moderate recall. Among our proposed variants, \emph{SCoT-ASR} produces fewer overlaps (41.1\%) and exhibits a bias toward silence, reflected in lower recall. This again confirms that ASR-only reasoning limits duplex responsiveness.

In contrast, \emph{SCoT-Response} produces substantially more overlaps (63.0\%), closer to natural dialogue rates, but with lower precision, suggesting that purely text-based reasoning favors responsiveness over cautious turn-taking. Finally, \emph{SCoT-Full} achieves the best balance, with the highest recall among all duplex systems, while maintaining more stable precision.
Overall, these results demonstrate that CoT reasoning, particularly when combining both ASR transcripts and text responses as intermediate targets, enables richer duplex behavior and yields interaction patterns more closely aligned with human-human dialogue.

\begin{table}[t]
    \centering
    \caption{Conversation Level Statistics on Switchboard.}
   {
    \begin{tabular}{lcc}
        \hline
        \textbf{Model} & \textbf{Emotion Rank (↓)}  & \textbf{Model Size (↓)} \\
        \hline
        Cascaded (Pre-train) & 5.75  & 3.4B\\
        Cascaded (Fine-tune) & 4.76 & 5.1B\\\midrule
        E2E  & 4.80 & \textbf{1.7B}  \\
        Multi turn CoT E2E &  3.35 &\textbf{1.7B}\\\midrule
        Duplex E2E & 5.60 &  \textbf{1.7B}\\\midrule
        \emph{SCoT-ASR} (proposed) & 6.54  &  \textbf{1.7B}\\
        \emph{SCoT-Response} (proposed) & \textbf{2.32} &  \textbf{1.7B}\\
        \emph{SCoT-Full} (proposed) & 2.90 &  \textbf{1.7B}\\
        \hline
    \end{tabular}
    }
    \label{tab:conversation_results}
\end{table}

\begin{table}[t]
    \centering
    \caption{Audio quality comparison.}
{
    \begin{tabular}{l|c}
        \hline
        \textbf{Model} & \textbf{UTMOS (↑)} \\
        \hline
        LJSpeech VITS (GT) & 4.19\hphantom{*} \\
        SpeechLM Pre-train (GT) & 2.08\hphantom{*} \\
        SpeechLM Finetune (GT)  & 2.21\hphantom{*} \\
        E2E   & 2.03\hphantom{*}\\\midrule
        Multi turn CoT E2E & 2.16\hphantom{*}\\\midrule
        Duplex E2E & 1.95\hphantom{*}\\ \midrule
        \emph{SCoT-ASR} (proposed)  & 1.89\hphantom{*} \\
        \emph{SCoT-Response} (proposed)  & 2.46\hphantom{*}\\
        \emph{SCoT-Full} (proposed) & 2.19\hphantom{*} \\
        \hline
    \end{tabular}
    }
    \label{tab:tts_results}
\end{table}

\subsection{Conversation Level Analysis}
Table~\ref{tab:conversation_results} presents a conversation-level analysis focusing on how well spoken dialogue systems preserve speaking styles such as emotion.
To enable fair comparison between turn-by-turn and duplex systems, we aggregate the speech output from each block within a turn and compute emotion similarity between the combined audio and the corresponding reference utterance. Our results show that \emph{SCoT-Response} achieves the strongest emotional alignment with human references (Emotion Rank = 2.32), substantially outperforming both other SCoT variants and all baseline systems, including turn-by-turn CoT E2E and cascaded models. \emph{SCoT} also performs competitively, while \emph{SCoT-ASR} again lags behind. Importantly, all SCoT models maintain high parameter efficiency, in contrast to larger cascaded systems, highlighting its potential in resource-constrained, on-device scenarios.
\subsection{Audio Quality}
Next, we evaluate audio quality performance in Table~\ref{tab:tts_results}. We observe that our proposed CoT formulation leads to a noticeable improvement in audio quality compared to the baseline Duplex E2E systems.\footnote{We can potentially improve audio quality by experimenting with higher quality speaker prompt as in \cite{arora2025cotsds}.} Further, \emph{SCoT-response} achieves highest audio quality among its variant, indicating that incorporating text response-based reasoning pathways not only improves semantic quality (Table~\ref{tab:text_response_results}) but also contributes to better speech naturalness.

\begin{table}[t]
    \centering
    \caption{Block size ablation for SCoT-Full models. Smaller blocks reduce latency (RTF) but hurt semantic quality, while a 2-second block offers a better balance.}
    {
    \begin{tabular}{l|cc}
        \hline
        \textbf{Block} & \multicolumn{2}{c}{SWBD} \\
        & \textbf{ROUGE-L (↑)}  & \textbf{RTF(↓)}  \\
        \hline
        1 sec & \hphantom{0}7.1 & 0.30\\
        2 sec & \textbf{15.8} & 0.85\\ 
        \hline
    \end{tabular}
    }
    \label{tab:block_size_results}
\end{table}
\subsection{On block size ablations} 
We trained an additional \emph{SCot-Full} model with a smaller block size of 1 second to analyze the trade-off between latency and response quality. While reducing the block size does lower latency as shown in Tab.~\ref{tab:block_size_results}, we observed a substantial degradation in semantic coherence: the model with 1-second blocks achieved a Perplexity of 161.2 and ROUGE-L of 7.06, compared to 55.2 and 20.3, respectively, for the 2-second block size (See Table 1). 
This degradation likely occurs because shorter blocks contain fewer words, limiting the contextual information available for intermediate reasoning steps such as ASR transcription and text response generation. This suggest that a 2-second block size offers a better balance between latency and response coherence.

\section{Conclusion}
\label{sec: conclusion}
We present SCoT, a novel Streaming Chain-of-Thought (CoT) Duplex architecture for spoken dialogue systems that integrates structured intermediate reasoning into a low-latency, blockwise streaming framework. By generating ASR transcripts and text responses at each step, SCoT enhances semantic coherence while preserving the real-time, full-duplex nature of time-multiplexing approaches. The SCoT variants reveal important insights into the trade-offs of different reasoning paths. \emph{SCoT-ASR}, which relies solely on transcript prediction, suffers from degraded semantic quality, confirming that ASR-only reasoning is insufficient in streaming interaction. \emph{SCoT-Response} emerges as the most effective variant overall—achieving the best semantic quality, strongest emotional alignment, and highest audio naturalness, while also maintaining very low latency.
\emph{SCoT-Full}, which incorporates both transcripts and text responses, achieves the strongest recall in overlap generation and stable duplex behavior, showing the benefit of explicitly modeling both reasoning stages, however with some additional latency.
Taken together, these results highlight SCoT as the \emph{first} systematic framework to analyze and optimize CoT reasoning paths in duplex dialogue systems. By balancing semantic quality, emotional expressivity, turn-taking behavior, and latency, SCoT sets a new standard for real-time, human-like conversational agents.

 A natural next step is to investigate adaptive block sizing strategies that dynamically balance latency against semantic coherence.
Finally, 
A deeper investigation into how different CoT paths influence system performance—and how their effectiveness varies across utterance types—could provide valuable insights into designing more robust and adaptive reasoning strategies.

\section*{ACKNOWLEDGMENT}
Experiments of this work used the Bridges2 system at PSC and Delta system at NCSA through allocations CIS210014 and IRI120008P from the Advanced Cyberinfrastructure Coordination Ecosystem: Services \& Support (ACCESS) program, supported by National Science Foundation grants \#2138259,\#:2138286, \#:2138307, \#:2137603, and \#:2138296.

\bibliographystyle{IEEEtran}
\bibliography{mybib}

\begin{thebibliography}{10}
\providecommand{\url}[1]{#1}
\csname url@samestyle\endcsname
\providecommand{\newblock}{\relax}
\providecommand{\bibinfo}[2]{#2}
\providecommand{\BIBentrySTDinterwordspacing}{\spaceskip=0pt\relax}
\providecommand{\BIBentryALTinterwordstretchfactor}{4}
\providecommand{\BIBentryALTinterwordspacing}{\spaceskip=\fontdimen2\font plus
\BIBentryALTinterwordstretchfactor\fontdimen3\font minus \fontdimen4\font\relax}
\providecommand{\BIBforeignlanguage}[2]{{%
\expandafter\ifx\csname l@#1\endcsname\relax
\typeout{** WARNING: IEEEtran.bst: No hyphenation pattern has been}%
\typeout{** loaded for the language `#1'. Using the pattern for}%
\typeout{** the default language instead.}%
\else
\language=\csname l@#1\endcsname
\fi
#2}}
\providecommand{\BIBdecl}{\relax}
\BIBdecl

\bibitem{jokinen2009spoken}
K.~Jokinen \emph{et~al.}, \emph{Spoken dialogue systems}.\hskip 1em plus 0.5em minus 0.4em\relax Morgan \& Claypool Publishers, 2009.

\bibitem{breazeal2008social}
C.~Breazeal \emph{et~al.}, ``Social robots that interact with people,'' \emph{Springer handbook of robotics}, pp. 1349--1369, 2008.

\bibitem{glass1999challenges}
J.~Glass, ``Challenges for spoken dialogue systems,'' in \emph{Proceedings of the 1999 IEEE ASRU Workshop}, vol. 696.\hskip 1em plus 0.5em minus 0.4em\relax MIT Laboratory for Computer Science Cambridge, 1999.

\bibitem{raux2005let}
A.~Raux \emph{et~al.}, ``Let's go public! taking a spoken dialog system to the real world,'' in \emph{Proc. Interspeech 2005}, 2005, pp. 885--888.

\bibitem{Qwen2.5-Omni}
J.~Xu \emph{et~al.}, ``Qwen2.5-omni technical report,'' \emph{arXiv preprint arXiv:2503.20215}, 2025.

\bibitem{pywebrtcvad}
\BIBentryALTinterwordspacing
J.~Wiseman, ``py-webrtcvad,'' 2024, accessed: 2024-12-10. [Online]. Available: \url{https://github.com/wiseman/py-webrtcvad}
\BIBentrySTDinterwordspacing

\bibitem{young2013pomdp}
S.~Young \emph{et~al.}, ``Pomdp-based statistical spoken dialog systems: A review,'' \emph{Proceedings of the IEEE}, vol. 101, pp. 1160--1179, 2013.

\bibitem{brown2020language}
T.~Brown \emph{et~al.}, ``Language models are few-shot learners,'' in \emph{Proc. NeurIPS}, 2020.

\bibitem{gpt4}
OpenAI \emph{et~al.}, ``{GPT-4} technical report,'' \emph{arXiv preprint arXiv:2303.08774}, 2024.

\bibitem{nguyen2024spirit}
T.~A. Nguyen \emph{et~al.}, ``{S}pi{R}it-{LM}: Interleaved spoken and written language model,'' \emph{Trans. ACL}, vol.~13, pp. 30--52, 2025.

\bibitem{jinchuan2024speechlm}
J.~Tian \emph{et~al.}, ``{ESPnet-SpeechLM}: An open speech language model toolkit,'' \emph{arXiv}, 2024.

\bibitem{arora2025landscapespokenlanguagemodels}
S.~Arora \emph{et~al.}, ``On the landscape of spoken language models: A comprehensive survey,'' 2025.

\bibitem{huggingface_speech_to_speech}
\BIBentryALTinterwordspacing
H.~Face, ``Speech-to-speech translation toolkit,'' 2024, accessed: 2024-12-10. [Online]. Available: \url{https://github.com/huggingface/speech-to-speech}
\BIBentrySTDinterwordspacing

\bibitem{xie2024miniomnilanguagemodelshear}
\BIBentryALTinterwordspacing
Z.~Xie \emph{et~al.}, ``Mini-omni: Language models can hear, talk while thinking in streaming,'' 2024. [Online]. Available: \url{https://arxiv.org/abs/2408.16725}
\BIBentrySTDinterwordspacing

\bibitem{ten2005temporal}
L.~Ten~Bosch \emph{et~al.}, ``On temporal aspects of turn taking in conversational dialogues,'' \emph{Speech Communication}, vol.~47, pp. 80--86, 2005.

\bibitem{arora2025talkingturnsbenchmarkingaudio}
S.~Arora \emph{et~al.}, ``Talking turns: Benchmarking audio foundation models on turn-taking dynamics,'' 2025.

\bibitem{TRP1}
\BIBentryALTinterwordspacing
A.~Gravano \emph{et~al.}, ``A corpus-based study of interruptions in spoken dialogue,'' in \emph{13th Annual Conference of the International Speech Communication Association, {INTERSPEECH} 2012, Portland, Oregon, USA, September 9-13, 2012}.\hskip 1em plus 0.5em minus 0.4em\relax {ISCA}, 2012, pp. 855--858. [Online]. Available: \url{https://doi.org/10.21437/Interspeech.2012-193}
\BIBentrySTDinterwordspacing

\bibitem{ma2024languagemodellistenspeaking}
\BIBentryALTinterwordspacing
Z.~Ma \emph{et~al.}, ``Language model can listen while speaking,'' 2024. [Online]. Available: \url{https://arxiv.org/abs/2408.02622}
\BIBentrySTDinterwordspacing

\bibitem{Dialog_GSLM}
\BIBentryALTinterwordspacing
T.~A. Nguyen \emph{et~al.}, ``Generative spoken dialogue language modeling,'' \emph{Trans. Assoc. Comput. Linguistics}, vol.~11, pp. 250--266, 2023. [Online]. Available: \url{https://doi.org/10.1162/tacl\_a\_00545}
\BIBentrySTDinterwordspacing

\bibitem{meng2024parrot}
Z.~Meng \emph{et~al.}, ``Parrot: Autoregressive spoken dialogue language modeling with decoder-only transformers,'' in \emph{NeurIPS Workshop AI-Driven Speech, Music, and Sound Generation}, 2024.

\bibitem{zhang2024turnbasedgameenablingrealtime}
X.~Zhang \emph{et~al.}, ``Beyond the turn-based game: Enabling real-time conversations with duplex models,'' in \emph{Proc. EMNLP}, 2024.

\bibitem{kyutai2024moshi}
\BIBentryALTinterwordspacing
A.~D\'efossez \emph{et~al.}, ``Moshi: a speech-text foundation model for real-time dialogue,'' Kyutai, Tech. Rep., 2024. [Online]. Available: \url{http://kyutai.org/Moshi.pdf}
\BIBentrySTDinterwordspacing

\bibitem{arora2025cotsds}
S.~Arora \emph{et~al.}, ``Chain-of-thought training for open e2e spoken dialogue systems,'' \emph{arXiv}, 2024.

\bibitem{zhang2023speechgpt}
D.~Zhang \emph{et~al.}, ``Speech{GPT}: Empowering large language models with intrinsic cross-modal conversational abilities,'' \emph{arXiv preprint arXiv:2305.11000}, 2023.

\bibitem{zhang2024omniflatten}
Q.~Zhang \emph{et~al.}, ``Omniflatten: An end-to-end {GPT} model for seamless voice conversation,'' \emph{arXiv preprint arXiv:2410.17799}, 2024.

\bibitem{wei2022chain}
J.~Wei \emph{et~al.}, ``Chain-of-thought prompting elicits reasoning in large language models,'' \emph{NeurIPS}, vol.~35, 2022.

\bibitem{zhang2022automatic}
Z.~Zhang \emph{et~al.}, ``Automatic chain of thought prompting in large language models,'' \emph{arXiv preprint arXiv:2210.03493}, 2022.

\bibitem{huang2024audiogpt}
R.~Huang \emph{et~al.}, ``Audio{GPT}: Understanding and generating speech, music, sound, and talking head,'' in \emph{Proceedings of the AAAI}, 2024.

\bibitem{zhang2024speechgpt}
D.~Zhang \emph{et~al.}, ``Speechgpt-gen: Scaling chain-of-information speech generation,'' \emph{arXiv preprint arXiv:2401.13527}, 2024.

\bibitem{castillo-lopez-etal-2025-survey}
\BIBentryALTinterwordspacing
G.~Castillo-L{\'o}pez \emph{et~al.}, ``A survey of recent advances on turn-taking modeling in spoken dialogue systems,'' in \emph{Proceedings of the 15th International Workshop on Spoken Dialogue Systems Technology}, M.~I. Torres \emph{et~al.}, Eds.\hskip 1em plus 0.5em minus 0.4em\relax Bilbao, Spain: Association for Computational Linguistics, May 2025, pp. 254--271. [Online]. Available: \url{https://aclanthology.org/2025.iwsds-1.27/}
\BIBentrySTDinterwordspacing

\bibitem{ekstedt2022voiceactivityprojectionselfsupervised}
\BIBentryALTinterwordspacing
E.~Ekstedt \emph{et~al.}, ``Voice activity projection: Self-supervised learning of turn-taking events,'' 2022. [Online]. Available: \url{https://arxiv.org/abs/2205.09812}
\BIBentrySTDinterwordspacing

\bibitem{EOT1}
R.~Masumura \emph{et~al.}, ``Neural dialogue context online end-of-turn detection,'' in \emph{SIGdial}.\hskip 1em plus 0.5em minus 0.4em\relax Association for Computational Linguistics, 2018, pp. 224--228.

\bibitem{turntakeingIS18}
M.~Roddy \emph{et~al.}, ``Investigating speech features for continuous turn-taking prediction using {LSTM}s,'' in \emph{Interspeech}, 2018.

\bibitem{skantze-2017-towards}
G.~Skantze, ``Towards a general, continuous model of turn-taking in spoken dialogue using {LSTM} recurrent neural networks,'' in \emph{Proc. {SIG}dial Meeting Disc. and Dial.}, 2017.

\bibitem{MEENA2014903}
R.~Meena \emph{et~al.}, ``Data-driven models for timing feedback responses in a map task dialogue system,'' \emph{Computer speech \& language}, vol.~28, pp. 903--922, 2014.

\bibitem{EOT2}
E.~Ekstedt \emph{et~al.}, ``Turn{GPT}: a transformer-based language model for predicting turn-taking in spoken dialog,'' in \emph{{EMNLP} 2020}, ser. Findings of {ACL}.\hskip 1em plus 0.5em minus 0.4em\relax Association for Computational Linguistics, 2020.

\bibitem{wang2024freezeomnismartlowlatency}
X.~Wang \emph{et~al.}, ``{Freeze-Omni}: A smart and low latency speech-to-speech dialogue model with frozen {LLM},'' \emph{arXiv preprint arXiv:2411.00774}, 2024.

\bibitem{xu2024enablingrealtimeconversationsminimal}
W.~Xu \emph{et~al.}, ``Enabling real-time conversations with minimal training costs,'' \emph{arXiv preprint arXiv:2409.11727}, 2024.

\bibitem{duplextimeNuerIPS24}
P.~Wang \emph{et~al.}, ``A full-duplex speech dialogue scheme based on large language models,'' in \emph{Proc. NeurIPS}, 2024.

\bibitem{veluri2024beyond}
B.~Veluri \emph{et~al.}, ``Beyond turn-based interfaces: Synchronous {LLM}s as full-duplex dialogue agents,'' \emph{arXiv preprint arXiv:2409.15594}, 2024.

\bibitem{Switchboard}
J.~Godfrey \emph{et~al.}, ``Switchboard: telephone speech corpus for research and development,'' in \emph{[Proceedings] ICASSP-92: 1992 IEEE International Conference on Acoustics, Speech, and Signal Processing}, vol.~1, 1992, pp. 517--520 vol.1.

\bibitem{cieri2004fisher}
C.~Cieri \emph{et~al.}, ``The fisher corpus: A resource for the next generations of speech-to-text.'' in \emph{LREC}, vol.~4, 2004, pp. 69--71.

\bibitem{lin2004rouge}
C.-Y. Lin, ``Rouge: A package for automatic evaluation of summaries,'' in \emph{Text summarization branches out}, 2004, pp. 74--81.

\bibitem{banerjee2005meteor}
S.~Banerjee \emph{et~al.}, ``Meteor: An automatic metric for mt evaluation with improved correlation with human judgments,'' in \emph{ACL workshop on intrinsic and extrinsic evaluation measures for machine translation and/or summarization}, 2005, pp. 65--72.

\bibitem{jelinek1977perplexity}
F.~Jelinek \emph{et~al.}, ``Perplexity—a measure of the difficulty of speech recognition tasks,'' \emph{The Journal of the Acoustical Society of America}, vol.~62, pp. S63--S63, 1977.

\bibitem{gpt-2}
A.~Radford \emph{et~al.}, ``Language models are unsupervised multitask learners,'' \emph{OpenAI}, 2019.

\bibitem{whisper}
A.~Radford \emph{et~al.}, ``Robust speech recognition via large-scale weak supervision,'' in \emph{Proc. ICML}, 2023.

\bibitem{shi2024versa}
J.~Shi \emph{et~al.}, ``{VERSA}: A versatile evaluation toolkit for speech, audio, and music,'' in \emph{"NAACL 2025"}, N.~Dziri \emph{et~al.}, Eds.\hskip 1em plus 0.5em minus 0.4em\relax Albuquerque, New Mexico: Association for Computational Linguistics, Apr. 2025.

\bibitem{saeki22c_interspeech}
T.~Saeki \emph{et~al.}, ``{UTMOS}: {UTokyo-SaruLab} system for voice{MOS} challenge 2022,'' in \emph{Interspeech}, 2022, pp. 4521--4525.

\bibitem{ji2024wavchatsurveyspokendialogue}
\BIBentryALTinterwordspacing
S.~Ji \emph{et~al.}, ``Wavchat: A survey of spoken dialogue models,'' 2024. [Online]. Available: \url{https://arxiv.org/abs/2411.13577}
\BIBentrySTDinterwordspacing

\bibitem{ma2023emotion2vec}
Z.~Ma \emph{et~al.}, ``emotion2vec: Self-supervised pre-training for speech emotion representation,'' \emph{Proc. ACL 2024 Findings}, 2024.

\bibitem{allal2025smollm2smolgoesbig}
\BIBentryALTinterwordspacing
L.~B. Allal \emph{et~al.}, ``Smollm2: When smol goes big -- data-centric training of a small language model,'' 2025. [Online]. Available: \url{https://arxiv.org/abs/2502.02737}
\BIBentrySTDinterwordspacing

\bibitem{hayashi2020espnet}
T.~Hayashi \emph{et~al.}, ``{ESPnet-TTS}: Unified, reproducible, and integratable open source end-to-end text-to-speech toolkit,'' in \emph{ICASSP}.\hskip 1em plus 0.5em minus 0.4em\relax IEEE, 2020, pp. 7654--7658.

\bibitem{owsm-v4}
Y.~Peng \emph{et~al.}, ``{OWSM} v4: Improving open whisper-style speech models via data scaling and cleaning,'' in \emph{Proceedings of the Annual Conference of the International Speech Communication Association (INTERSPEECH)}, 2025.

\bibitem{espnet}
S.~Watanabe \emph{et~al.}, ``{ESPnet}: End-to-end speech processing toolkit,'' in \emph{Proceedings of Interspeech}, 2018, pp. 2207--2211.

\bibitem{ESPnet-SLU}
S.~Arora \emph{et~al.}, ``{ESPnet}-{SLU}: Advancing spoken language understanding through espnet,'' in \emph{ICASSP}.\hskip 1em plus 0.5em minus 0.4em\relax IEEE, 2022, pp. 7167--7171.

\bibitem{arora2025espnet}
S.~Arora \emph{et~al.}, ``Espnet-sds: Unified toolkit and demo for spoken dialogue systems,'' in \emph{NAACL 2025}, 2025, pp. 248--259.

\bibitem{musicgen}
J.~Copet \emph{et~al.}, ``Simple and controllable music generation,'' \emph{Advances in Neural Information Processing Systems}, vol.~36, 2024.

\bibitem{shi2024espnet}
J.~Shi \emph{et~al.}, ``{ESPnet-Codec}: Comprehensive training and evaluation of neural codecs for audio, music, and speech,'' \emph{arXiv preprint arXiv:2409.15897}, 2024.

\bibitem{chen2024towards}
W.~Chen \emph{et~al.}, ``Towards robust speech representation learning for thousands of languages,'' \emph{arXiv preprint arXiv:2407.00837}, 2024.

\end{thebibliography}

\vfill\pagebreak

\end{document}